\documentclass[conference]{IEEEtran}
\IEEEoverridecommandlockouts
\usepackage{cite}
\usepackage{amsmath,amssymb,amsfonts}
\usepackage{algorithmic}
\usepackage{graphicx}
\usepackage{textcomp}
\usepackage{xcolor}
\def\BibTeX{{\rm B\kern-.05em{\sc i\kern-.025em b}\kern-.08em
    T\kern-.1667em\lower.7ex\hbox{E}\kern-.125emX}}
\begin{document}

\title{Conversational Context Classification: A Representation Engineering Approach\\
{\footnotesize \textsuperscript{ }}
\thanks{Identify applicable funding agency here. If none, delete this.}
}

\author{\IEEEauthorblockN{ Jonathan Pan}
\IEEEauthorblockA{\textit{Nanyang Technological University} \\
Singapore \\
JonathanPan@ntu.edu.sg}

}

\maketitle

\begin{abstract}
The increasing prevalence of Large Language Models (LLMs) demands effective safeguards for their operation, particularly concerning their tendency to generate “out-of-context” responses. A key challenge is accurately detecting when LLMs stray from expected conversational norms, manifesting as topic shifts, factual inaccuracies, or outright hallucinations. Traditional anomaly detection struggles to directly apply within contextual semantics. This paper outlines our experiment in exploring the use of Representation Engineering (RepE) and One-Class Support Vector Machine (OCSVM) to identify subspaces within the internal states of LLMs that represent a specific context. By training OCSVM on in-context examples, we establish a robust boundary within the LLM’s hidden state latent space. We evaluated our study with two open source LLMs – Llama and Qwen models in specific contextual domain. Our approach involved identifying the optimal layers within the LLM’s internal state subspaces that strongly associates with the context of interest. Our evaluation results showed promising results in identifying the subspace for a specific context. Aside from being useful with detecting in or out of context conversation threads for AI safety, this research work contributes to the study of better interpreting LLMs. 
\end{abstract}


\section{Introduction}
The emergence of Large Language Models (LLMs) represents a paradigm shift in human-computer interaction, dramatically altering the landscape of communication and information processing. These models, capable of generating remarkably coherent and contextually relevant text, are rapidly permeating a diverse range of applications – from sophisticated chatbots and automated content creation to complex problem-solving and the conduct of research work. However, this transformative potential is inextricably linked to significant challenges, primarily stemming from the “black-box” nature of LLMs and the emergent, often unpredictable, behaviors they exhibit. A particularly critical concern is the tendency of LLMs to generate “out-of-context” responses, manifesting as abrupt topic drifts, demonstrable factual inaccuracies (commonly referred to as “hallucinations”), logical inconsistencies, or a fundamental failure to adhere to established conversational norms. Such deviations not only frustrate users but also pose serious risks, potentially disseminating misinformation and ultimately eroding trust in AI systems as reliable tools. 

Ensuring that LLMs maintain conversational coherence and remain firmly “in-context” is therefore paramount for their safe and effective deployment across a spectrum of high-stakes applications. Traditional methods for evaluating conversational quality frequently rely on rule-based systems or, more commonly, post-hoc human evaluation – approaches that are often prohibitively costly, exceptionally time-consuming, and inherently reactive. 

While the opacity of the inner workings of LLM limits interpretation, there are signals being generated when a LLM processes an input prompt. These signals are high dimensional. A new form of analysis technique called Representation Engineering (RepE) [4] recently emerged from the study of LLMs. This involves analyzing and manipulating the internal representations of these AI models to understand their behavior which can be used to proactively detect undesirable conversational patterns.

This paper outlines our experimental study of applying RepE with One-Class Support Vector Machines (OCSVM) on the hidden states of LLMs to identify subspaces that represents a specific context. Our research hypothesis is that “in-context” conversational turns would occupy a discernible, compact region within the LLM’s high-dimensional hidden state space, and that “out-of-context” turns will invariably fall outside this boundary, thus being readily detectable as anomalies. By training a OCSVM exclusively on examples of normal, in-context dialogue, we establish a robust and informative boundary.

Our contributions are multifaceted: we propose and implemented a One-Class SVM based framework for detecting in/out-of-context conversations by analyzing the hidden states of a pre-trained LLM; we conducted a comprehensive empirical evaluation of this method utilizing Meta’s Llama 3.2 3B and Qwen’s Qwen2.5 3B models with specialized  conversational domain, demonstrating its practical applicability; we performed rigorous ablation studies to assess the impact of different hidden layers on detection performance, providing valuable insights into optimal configurations; and finally, we present a detailed error analysis, meticulously categorizing common failure modes and discuss their implications for understanding LLM internal context representation and refining future detection strategies.

In the next section, we will first cover related work on this topic. We then describe the empirical method used in this work which is followed by the experimental setup and an analysis of our results. We conclude our research with a discussion about future research directions
.

\section{Related Work}

The problem of in/out-of-context detection in Large Language Models (LLMs) is situated within a complex landscape of research, drawing connections to several interconnected areas. This section outlines the key related work, highlighting the foundations and context for our proposed approach.

\subsection{LLM Safety and Alignment}

A significant body of research focuses on ensuring the safe and reliable operation of LLMs. The field of LLM safety [1] addresses concerns regarding harmful, biased, or unaligned outputs, with “context drift” and hallucination [2] representing major challenges to their trustworthiness. Our work builds upon this foundation by providing a method for proactively detecting these reliability failures within conversational contexts. 

\subsection{Representation Engineering}

Representation Engineering (RepE) is a paradigm in AI interpretability and safety that focuses on direct analysis and modification of a latent subspace based on a transformer based LLM model [3] to observe or steer behavior, without the need to retrain the entire network [4]. Unlike classical fine-tuning approaches that involve modification of weights, representation engineering identifies and intervenes on semantically meaningful subspaces of activations. Recent work with RepE involved editing factual associations [5, 6] in language models by locating and adjusting linear directions in hidden layers. 

\subsection{Anomaly and Out-of-Distribution Detection}

This broad field encompasses techniques to identify data points that deviate significantly from the norm [5]. Several anomaly detection algorithms have been developed, including: One-Class Support Vector Machines (OCSVM) [6], a prominent unsupervised/semi-supervised algorithm that learns a decision boundary encompassing a single class of “normal” data, classifying all other points as outliers or novelties. Traditional anomaly detection metrics such as statistical tests, distance-based approaches (e.g., k-NN), density-based methods (e.g., LOF), and reconstruction-based methods (e.g., Autoencoders) [7]are other forms of detectors. 

\subsection{Anomaly Detection in NLP/LLM}

The application of anomaly detection techniques to textual data, particularly in the context of LLMs, has gained traction. Early work explored anomaly detection for spam or misinformation detection [8]. More recently, researchers have used these methods to analyze LLM output or their internal representations for various purposes, including identifying out-of-distribution conversations or detecting adversarial attacks [9]. Several studies have benchmarked different embedding-based anomaly detection algorithms for text, including OCSVM [10].

Our work distinguishes itself through a focused approach: (i) we empirically evaluate OCSVM on the hidden states of large, modern LLMs, providing insights specific to their complex internal representations; and (ii) we conduct an ablative analysis across different hidden layers to understand the representational hierarchy relevant to context, ultimately contributing a more nuanced understanding of LLM context representation
.

\section{Method}
Our proposed framework for detecting in/out-of-context conversations using One-Class Support Vector Machines (OCSVM) is applied to the hidden state subspaces of the LLM. It operates in three distinct phases: Calibration, Threshold Tuning, and Evaluation. This structured approach ensures robust and reliable detection of novel conversational turns.

\subsection{LLM Hidden State Extraction}\label
We utilized pre-trained open source Large Language Model (LLM) models. Our approach involved using the hidden state vectors within an "in-context" conversational turn that serves as the features space for the OCSVM. The OCSVM is then trained on a dataset of in context data samples. During training, the OCSVM algorithm learns a compact region (high dimensional subspace) that encapsulates these "normal" hidden states. To access these internal representations, we load the model with \texttt{output\_hidden\_states=True} in the Hugging Face Transformers library. This setting instructs the model to output the hidden states corresponding to each token at every layer of its transformer architecture.

\subsection{Context Detection Training}
The detection training process is structured into three phases – Calibration, Threshold Tuning and Evaluation Phases. The Calibration Phase involved a set of calibration examples (strictly in-context conversations). For each example, all token hidden states from the layers are extracted. These collected layer wise hidden states form the training dataset for the OCSVM. The OCSVM is fitted to the dataset layer wise, learning the decision boundary for ‘normal’ in-context representations.

The Threshold Tuning Phase involved using a data set containing a mix of labeled in-context and out-of-context examples. For each example in this set, its hidden states are passed through the trained OCSVM to obtain the average decision function scores. Various possible thresholds (derived from the range of calculated scores) are evaluated. The optimal threshold is selected by maximizing the overall accuracy (F1 score for imbalanced datasets) of predicting in-context vs out-of-context turns on this tuning set.

The final phase is the Evaluation Phase that involved a testing dataset, distinct from both calibration and tuning data, used for the final performance assessment. For each turn in this set, the trained OCSVM processes its hidden states and calculates an average decision function score. If this average score falls below the optimal threshold determined in the tuning phase, the turn is classified as "out-of-context." The model's predictions are compared against the true labels to calculate various evaluation metrics. 

\section{Experimental Setup}
In this section, we evaluated the effectiveness of our One-Class Support Vector Machine (OCSVM)-based anomaly detection framework, we constructed a specialized conversational dataset and implemented a controlled experimental setup. This section details the dataset construction, implementation details, and evaluation metrics employed to assess the detector’s performance.

\subsection{Dataset Construction}\label
The core of our evaluation relies on a carefully curated dataset designed to clearly differentiate between in-context and out-of-context conversational turns within the domain of AI/ML. This domain was selected due to its inherent contrast to the selected topic (AI/ML and AI safety), while extraneous topics represent anomalies. The dataset is partitioned into three distinct sets, each serving a specific purpose in our evaluation. The first set is the Calibration Set of 20 samples. It comprised of statements directly related to topics. It serves as the training data for the OCSVM, effectively teaching it the characteristics of the ‘normal’ in-context conversations within this domain. Examples include prompts such as a short prompt on AI/ML - ‘Discuss the challenges of training very deep neural networks.’ and long prompt on AI Safety – ‘Interpretability, also known as explainable AI (XAI), is crucial for safety. If we cannot understand why an AI model makes a particular decision, it is difficult to trust it, debug it, or ensure it is not operating on flawed or malicious reasoning. Techniques like feature visualization and saliency maps aim to make these 'black boxes' more transparent. 

For the Threshold Tuning and Evaluation phases, 20 samples were generated and used for each phase. Both datasets contain 10 in-context (Negative) and 10 out-of-context (Positive) samples. Example of the in-context is ‘What are the components of an LSTM cell?’ while out-of-context example is ‘What is the recipe for a classic margarita?’ that are deliberately defined to challenge the detector. Both these samples are labelled to facilitate evaluation
.

As our experiment is exploratory endeavour, hence our dataset size was kept small. Additionally, we needed the conversational samples to be strongly correlated to train the One-Class SVM to learn the decision boundary. The small sample sizes is applicable especially when the number of turns to conversational chats may be short.

\subsection{Implementation Details}
The experimental implementation involved two open source models (Llama3.2-3B and Qwen2.5-3B). The hidden states were extracted during inferences. The libraries used on Hugging Face Transformers for seamless interaction with the open source models, scikit-learn for the OneClassSVM model, and NumPy for efficient numerical operations. The hidden states from all layers of the LLM were used. As the hidden states of each layer are generated from feed forward neurons of the transformer decoder, we chose to use the last token that contains the decoder layer as it represented the most likely candidate for the last layer of the LLM that would generate the response to the given input. 

\subsection{Evaluation Metrics}
To thoroughly assess the performance of the OCSVM detector, we computed a suite of relevant evaluation metrics on the testing data:

\begin{itemize}
\item Accuracy: The overall percentage of correctly classified instances (both in-context and out-of-context).
\item Precision (for Out-of-Context class): Calculated as (True Positives + False Positives) / True Positives, measuring the proportion of correctly identified out-of-context turns among all instances flagged as anomalous.
\item Recall (for Out-of-Context class): Calculated as (True Positives + False Negatives) / True Positives, measuring the proportion of actual out-of-context turns that were correctly detected.
\item F1-score (for Out-of-Context class): The harmonic mean of Precision and Recall, providing a balanced measure, particularly important for imbalanced datasets where anomalies are often rare.
\item AUROC (Area Under Receiver Operating Characteristic Curve): Measures the model’s ability to discriminate between in-context and out-of-context turns across a range of decision thresholds.
\item AUPRC (Area Under Precision-Recall Curve): Specifically designed for anomaly detection, this metric focuses on the performance of the positive (minority) class, providing a robust assessment of the detector’s ability to identify rare anomalies
. 
\end{itemize}
\section{Results}
\subsection{Overall Performance}\label
The OCSVM model, trained on the high-dimensional hidden states of the both models, demonstrated strong performance in detecting out-of-context conversations. Across the layers, there are varying performances. There was a number of layers that had very good evaluation results.

\begin{figure}[h]
    \includegraphics[width=90mm,keepaspectratio]{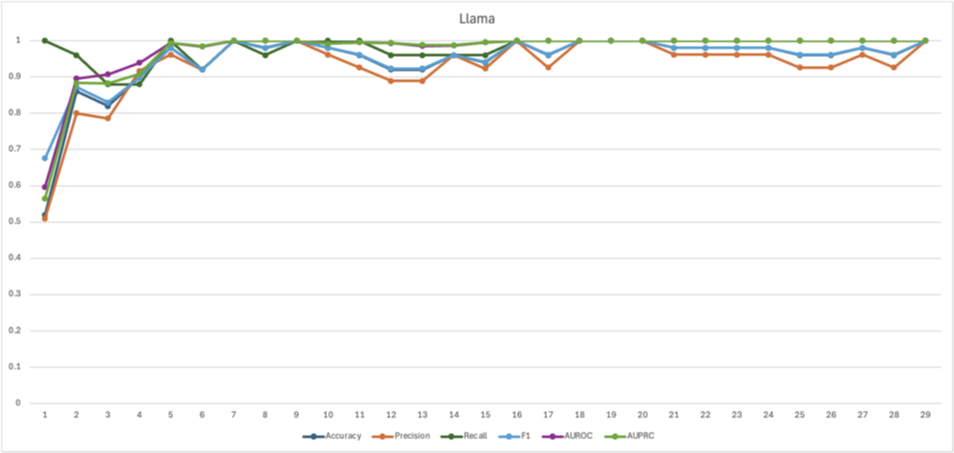}
Figure 1: Evaluation Performance of Llama on topic of ‘AI/ML’
\end{figure}

\begin{figure}[h]
    \includegraphics[width=90mm,keepaspectratio]{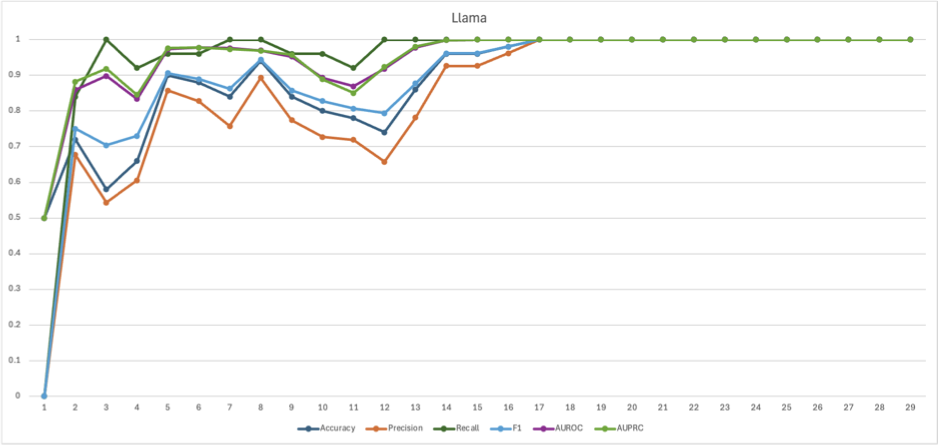}
Figure 2: Evaluation Performance of Llama on topic of ‘AI Safety’
\end{figure}

\begin{figure}[h]
    \includegraphics[width=90mm,keepaspectratio]{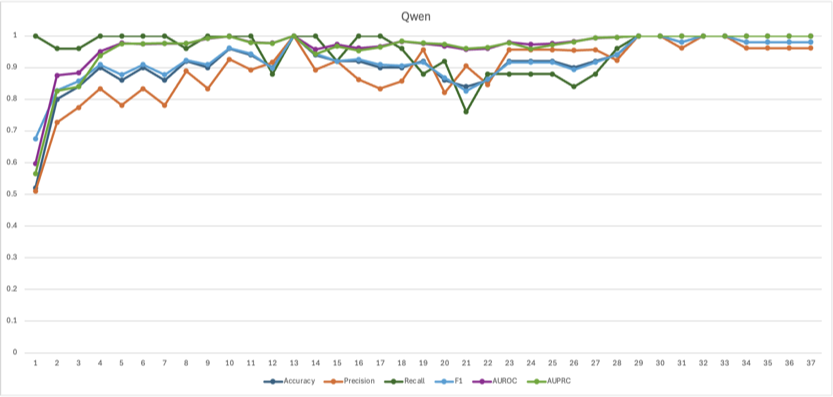}
Figure 3: Evaluation Performance of Qwen on topic of ‘AI/ML’
\end{figure}

\begin{figure}[h]
    \includegraphics[width=90mm,keepaspectratio]{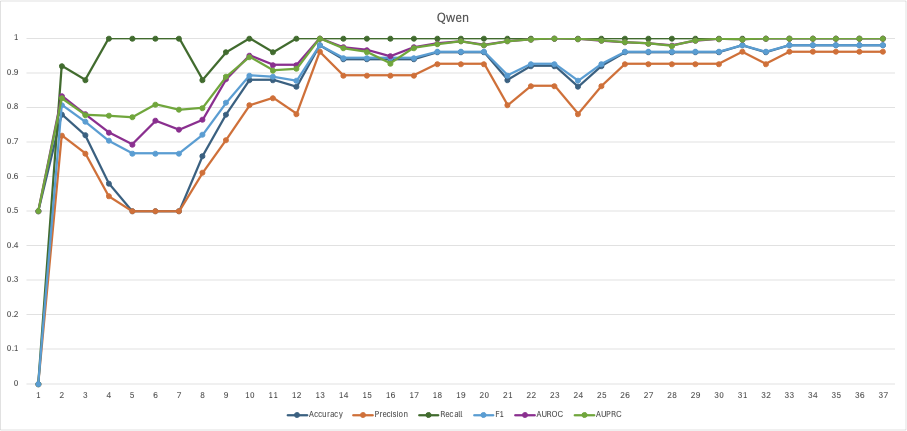}
Figure 4: Evaluation Performance of Qwen on topic of ‘AI Safety’
\end{figure}

These results (Figure 1 to 4) indicate that OCSVM, utilizing LLM hidden states, can effectively distinguish between in-context and novel out-of-context conversational turns. The high F1-score and AUROC suggest good overall discriminative evaluation performances.

\subsection{Visualization of Context Points}
After we evaluated the detector's performance using hidden states from different layers of both models, we studied the visual plot of the in-context and out-of-context points, when mapped onto a two-dimensional plane using Principal Component Analysis (PCA) of the hidden states from the layer that gave the best evaluation performance. For both models, the visual plots (Figure 5 and 6) show a clear separation of the point types based on one of high performance OCSVM classifier with its corresponding layers.

\begin{figure}[h]
    \centering
    \includegraphics[width=90mm,keepaspectratio]{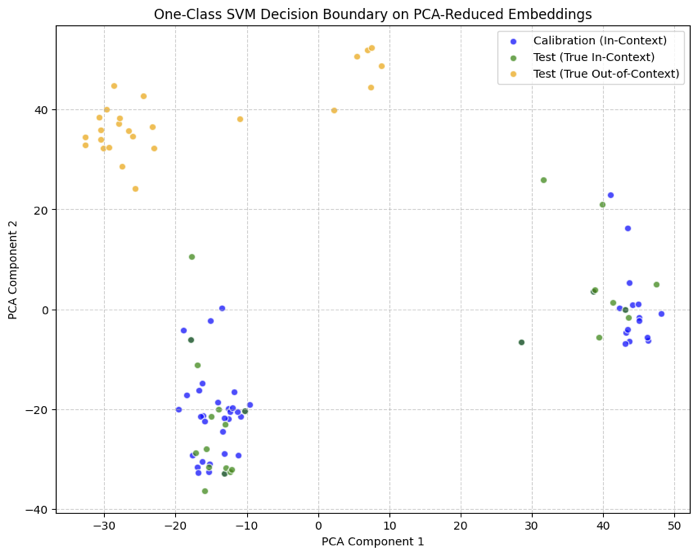}
Figure 5: PCA plot for Llama
\end{figure}

\begin{figure}[h]
    \centering
    \includegraphics[width=90mm,keepaspectratio]{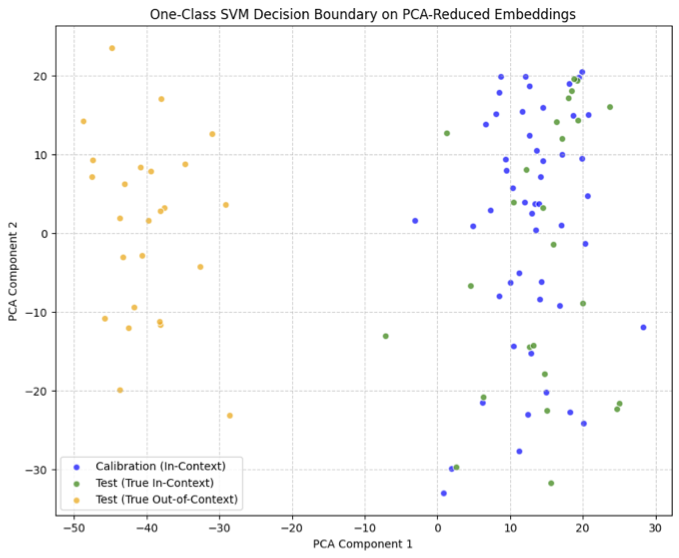}
Figure 6: PCA Plot for Qwen
\end{figure}

\section{Discussion}
Our empirical results establish that One-Class SVM can be applied to LLM’s hidden states as a method for assessing context. The consistently superior performance observed when utilizing hidden states underscores the critical role of latent subspaces within LLM in encoding semantically rich and contextually informed representations – a key factor for effective context analysis. Additionally, the significant impact of the layer selection within the LLM demonstrates the importance of choosing layers that effectively capture the appropriate level of semantic context with the specific conversational domain. This emphasizes the need for careful consideration when adapting the approach to different contexts. 

Finally, the ability to visualize the anomaly within the LLM’s latent space offers a degree of interpret-ability that enables further investigation using representation engineering or mechanistic interpret-ability techniques to fully understand the underlying causes. Moving forward, addressing the limitations through integration with these complementary methods will be crucial in realizing the full potential of this approach in robust and reliable conversational AI systems
.

\section{CONCLUSION AND FUTURE DIRECTIONS}
This exploratory research is a preliminary endeavour to apply Representation Engineering (RepE) approach to classify in or out of context conversational turns. We applied the One-Class Support Vector Machines (OCSVM) on Large Language Model (LLM) hidden states. Our preliminary assessment is that it is a robust and effective method for detecting out-of-context conversational turns. Our study, which models the “in-context” conversational space as a statistical distribution within the LLM’s latent representations, demonstrates a practical and adaptable framework for enhancing the safety, reliability, and controllability of conversational AI systems. The results highlight the value of leveraging the inherent hidden states of LLMs to proactively manage conversational drift and maintain coherence.

There are several areas that we could further explore with this exploratory research work. Firstly, integrating OCSVM with Sparse Autoencoders (SAEs) offers a promising avenue for enhanced interpretability. Training SAEs on LLM hidden states would generate sparse, interpretable features, allowing us to understand what kind of novelty is being detected – for example, identifying the specific activation of irrelevant topic features. This would move beyond simply flagging an anomaly to providing actionable insights. Combining OCSVM detection with causal tracing or activation patching would allow us to pinpoint the specific internal LLM components that causally contributed to the anomalous hidden state, providing granular explanations for the behavior and facilitating targeted interventions. Another is to extend the approach to multi-modal LLMs, incorporating data streams beyond just text, represents a significant opportunity. The ability to analyze conversations incorporating images, audio, or other data sources would dramatically broaden the scope of anomaly detection. 

This work lays a foundation for data-driven, interpretable anomaly detection in LLMs, and we believe it will significantly contribute to the advancement of safe and reliable conversational AI systems.

\section*{References}

\noindent[1]	Amodei, D., Olah, C., Steinhardt, J., Christiano, P., Schulman, J., \& Mané, D., Concrete problems in AI safety. arXiv preprint arXiv:1606.06565, 2016.

\noindent[2]	Huang, L., Yu, W., Ma, W., Zhong, W., Feng, Z., Wang, H., Chen, Q., Peng, W., Feng, X., Qin, B \& Liu, T., Survey of hallucination in large language models: Principles, Taxonomy, Challenges, and Open Questions. ACM Transactions on Information Systems, Vol. 43, Issue 2, 1-55, 2025.

\noindent[3]	Vaswani, A., Shazeer, N., Parmar, N., Uszkoreit, J., Jones, L., Gomez, A. N., Kaiser, L. \& Polosukhin, I., Attention is all you need. 31st Conference on Neural Information Processing Systems (NIPS 2017), 2017.

\noindent[4]	Zou, A., Phan, L., Chen, S., Campbell, J., Guo, P., Ren, R., Pan, A., Yin, X., Mazeika, M., Dombrowski, A.K., Goel, S., Li, N., Byun, M.J., Wang, Z., Mallen, A., Basart, S., Koyejo, S., Song, D., Fredrikson, M., Kolter, J.Z. \& Hendrycks, D., Representation Engineering: A Top-Down Approach to AI Transparency, arxiv:2310.01405, 2023.

\noindent[5]	Chandola, V., Banerjee, A., \& Kumar, V., Anomaly detection: A survey. ACM computing surveys (CSUR), Vol. 41, Issue 3, 1-58, 2009.

\noindent[6]	Schölkopf, B., Platt, J. C., Shawe-Taylor, J., Smola, A. J., \& Williamson, R. C. (2001). Estimating the support of a high-dimensional distribution. Neural computation, Vol. 13, Issue 7, 2001.

\noindent[7]	Ruff, L., Vandermeulen, R., Goernitz, N., Deecke, F., Siddiqui, S.A., Binder, A., Muller, E. \& Kloft, A., Deep One-Class Classification. Proceedings of the 35th International Conference on Machine Learning. PMLR 80:4393-4402, 2018.

\noindent[8]	Kaddoura, S., Chandrasekaran, G., Popescu, D.E. \& Duraisamy, J.H., A systematic literature review on spam content detection and classification. PeerJ Comput Sci., 2022. 

\noindent[9]	Maimon, G., \& Rokach, L, A Universal Adversarial Policy for Text Classifiers, arxiv:2206.09458, 2022.

\noindent[10]	Cao, Y., Yang, S., Li, C., Xiang, H., Qi, L., Liu, B., Li, R. \& Liu, M. (2025). TAD-Bench: A Comprehensive Benchmark for Embedding-Based Text Anomaly Detection. arXiv preprint arXiv:2501.11960.

\end{document}